\documentclass[a4paper,conference]{IEEEtran}

\usepackage[dvipsnames, table]{xcolor}         
\usepackage{hyperref}
\usepackage{url}
\usepackage[pdftex]{graphicx}
\usepackage[percent]{overpic}
\usepackage{amsfonts}  
\usepackage{booktabs}  
\usepackage{multirow}
\usepackage[caption=false,font=footnotesize]{subfig}
\usepackage{cite}
\usepackage{url}
\usepackage{bm}
\usepackage[separate-uncertainty]{siunitx}
\usepackage[percent]{overpic}
\usepackage{algorithmic}

\graphicspath{{images/}}

\definecolor{good}{HTML}{cccccc}

\begin{document}
%
\title{Autoencoder Attractors for Uncertainty Estimation}

\author{\IEEEauthorblockN{Steve Dias Da Cruz\IEEEauthorrefmark{1}\IEEEauthorrefmark{2}\IEEEauthorrefmark{3},
Bertram Taetz\IEEEauthorrefmark{3},
Thomas Stifter\IEEEauthorrefmark{1},
Didier Stricker\IEEEauthorrefmark{2}\IEEEauthorrefmark{3}}
\IEEEauthorblockA{\IEEEauthorrefmark{1}IEE S.A., \IEEEauthorrefmark{2}University of Kaiserslautern, \IEEEauthorrefmark{3}German Research Center for Artificial Intelligence (DFKI)}
\IEEEauthorblockA{Email: steve.dias-da-cruz@iee.lu, bertram.taetz@dfki.de, thomas.stifter@iee.lu, didier.stricker@dfki.de}}

\maketitle

\begin{abstract}
The reliability assessment of a machine learning model's prediction is an important quantity for the deployment in safety critical applications. Not only can it be used to detect novel sceneries, either as out-of-distribution or anomaly sample, but it also helps to determine deficiencies in the training data distribution. A lot of promising research directions have either proposed traditional methods like Gaussian processes or extended deep learning based approaches, for example, by interpreting them from a Bayesian point of view. In this work we propose a novel approach for uncertainty estimation based on autoencoder models: The recursive application of a previously trained autoencoder model can be interpreted as a dynamical system storing training examples as attractors. While input images close to known samples will converge to the same or similar attractor, input samples containing unknown features are unstable and converge to different training samples by potentially removing or changing characteristic features. The use of dropout during training and inference leads to a family of similar dynamical systems, each one being robust on samples close to the training distribution but unstable on new features. Either the model reliably removes these features or the resulting instability can be exploited to detect problematic input samples. We evaluate our approach on several dataset combinations as well as on an industrial application for occupant classification in the vehicle interior for which we additionally release a new synthetic dataset.     
\end{abstract}


%
\IEEEpeerreviewmaketitle

\section{Introduction}
Assessing the reliability of machine learning models' predictions is an important challenge for the deployment and applicability of statistical methods. This additional information allows the possibility to detect novel and exotic sceneries during the lifetime of a deployed model on which the model's predictions trustability can be determined. This knowledge also gives hints whether the collected training data needs to be extended or modified, e.g. in the case of active learning \cite{gal2017deep} and continuous learning \cite{kading2016fine}. Recent activities investigated the possibility for estimating the uncertainty in the case of deep learning based methods \cite{damianou2013deep, kupinski2003ideal, louizos2017multiplicative, amini2020deep}. Monte Carlo (MC) dropout, i.e. using dropout during training and enabling the latter during inference for multiple runs, has been shown to produce good uncertainty quantification \cite{gal2016dropout} on several tasks while limiting the additional overheat during training and inference.

\begin{figure}
  \centering
  \subfloat[Reconstructions of a same test sample from $D_{in}$ (GTSRB)]{
    \begin{overpic}[height=3.2cm]{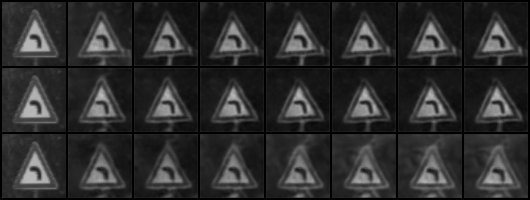}
      \put(3,35){\scriptsize{\textcolor{white}{Input}}}
      \put(16,35){\scriptsize{\textcolor{white}{Iter 1}}}
      \put(28,35){\scriptsize{\textcolor{white}{Iter 2}}}
      \put(41,35){\scriptsize{\textcolor{white}{Iter 3}}}
      \put(53,35){\scriptsize{\textcolor{white}{Iter 4}}}
      \put(65.5,35){\scriptsize{\textcolor{white}{Iter 5}}}
      \put(78,35){\scriptsize{\textcolor{white}{Iter 6}}}
      \put(90.5,35){\scriptsize{\textcolor{white}{Iter 7}}}
    \end{overpic}
  }
  \\
  \subfloat[Reconstructions of a same OOD sample from $D_{out}$ (SVHN)]{
    \begin{overpic}[height=3.2cm]{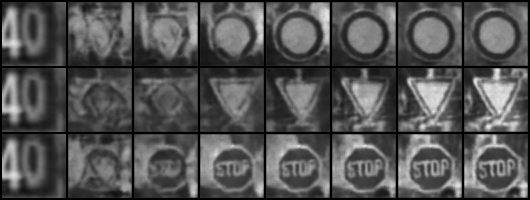}
      \put(3,35){\scriptsize{\textcolor{white}{Input}}}
      \put(16,35){\scriptsize{\textcolor{white}{Iter 1}}}
      \put(28,35){\scriptsize{\textcolor{white}{Iter 2}}}
      \put(41,35){\scriptsize{\textcolor{white}{Iter 3}}}
      \put(53,35){\scriptsize{\textcolor{white}{Iter 4}}}
      \put(65.5,35){\scriptsize{\textcolor{white}{Iter 5}}}
      \put(78,35){\scriptsize{\textcolor{white}{Iter 6}}}
      \put(90.5,35){\scriptsize{\textcolor{white}{Iter 7}}}
    \end{overpic}  
  }
  \caption{Multiple recursive reconstructions (from left to right) of identical samples (first column) from $D_{in}$ and $D_{out}$ by our novel model. Notice the evolution in the reconstructions over each iterative step for the OOD sample.}
  \label{fig:teaser}
\end{figure}
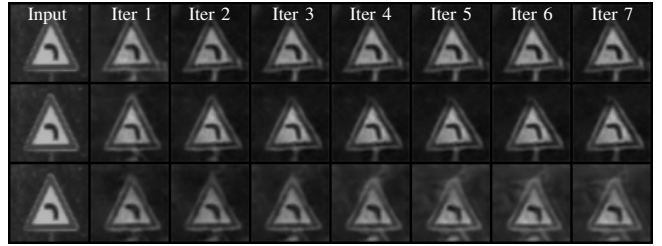
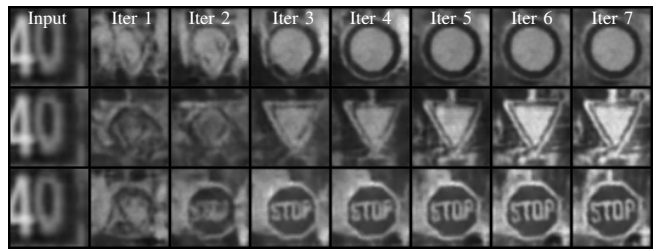

It has been shown that recursive applications of autoencoders, which are trained under the standard training regime, can be viewed as a dynamical system \cite{radhakrishnan2020overparameterized}. In mathematics \cite{strogatz2000}, the analysis of fixed points, attractors and their basins of attraction are important tools to analyze and understand dynamical systems and their behavior. This iterative process can be viewed as associative memory \cite{radhakrishnan2020overparameterized} to retrieve perturbed training samples, but the models need to be trained long enough to ensure that the training samples become attractors. To the best of our knowledge, the recursive application of autoencoders and their attractors have not been investigated in view of generalization and uncertainty estimation. 

Our contribution consists of the extension of the recursive application of autoencoder models, thus dynamical systems and attractors, in view of generalization capacities. We combine this strategy with MC dropout and we exploit characteristics of both design choices to determine whether new input samples are close or far from the training distribution by analyzing the behavior of multiple inferences, as shown in Fig. \ref{fig:teaser}: the test sample is converging to a similar attractor, while the out-of-distribution (OOD) sample converges to different attractors of different classes. We show that uncertainty estimation is improved compared to vanilla MC dropout and deep ensemble models across three metrics and in view of the entropy distribution. Our ablation study shows that the recursive application is key to the success of our approach. Our analysis is performed on several commonly used OOD dataset combinations as well as on an industrial application. We consider occupant classification in the vehicle interior and highlight some additional challenges. To this end we release a synthetic dataset for uncertainty estimation which will extend the existing SVIRO \cite{DiasDaCruz2020SVIRO} dataset for occupant classification.

\section{Related Works}

\noindent \textbf{Attractors:} There are several types of models achieving associative memory, e.g. discrete and continuous Hopfield Networks \cite{ramsauer2020hopfield, krotov2016dense, hopfield1982neural} and Predictive Coding \cite{salvatori2021associative}. The former needs an energy function to be defined, while the latter is biologically inspired. However, we focus on associative memory achieved by the recursive application of autoencoder models \cite{radhakrishnan2020overparameterized}, previously trained with gradient descent, due to their elegant simplicity and analogy to dynamical systems, which has been investigated extensively in mathematics and physics \cite{strogatz2000}. While a few works investigate properties of this model design \cite{jiang2020associative, radhakrishnan2020overparameterized, radhakrishnan2019memorization}, only one \cite{hadjahmadi2019robust} considers attractors for classification and uncertainty estimation. However, the latter adopts this only for speech recognition with respect to noise robustness and combines it with a hidden Markov model. We, on the contrary, apply this methodology to computer vision and assess the robustness against novel classes and unseen samples from either new datasets or the test distribution.  

\noindent \textbf{Uncertainty estimation:} A lot of research \cite{abdar2021review} is focusing on estimating the uncertainty of a model's prediction regarding OOD or anomaly detection, both of which are tightly related. However, only a few works consider the use of autoencoder models for assessing uncertainty: Autoencoders can be combined with normalizing flow \cite{bohm2020probabilistic}, refactor ideas from compressed sensing \cite{grover2019uncertainty} or use properties of Variational Autoencoders \cite{ran2021detecting, xiao2020likelihood}. More commonly, autoencoders are used for non image based datasets \cite{vartouni2018anomaly, xu2018unsupervised, oh2018residual}. Other deep learning approaches are based on evidential learning \cite{NEURIPS2018_a981f2b7, amini2020deep}, Bayesian methods \cite{mackay1995probable}, Variational Bayes \cite{blundell2015weight} or on Hamiltonian Monte-Carlo \cite{chen2014stochastic}. Also non deep-learning approaches have shown significant success, but are less scalable, as for example Gaussian Processes \cite{rasmussen2003gaussian} or approaches based on support vector machines \cite{noori2015uncertainty}. Since our approach borrows ideas from MC dropout \cite{gal2016dropout}, we limit our comparison against the latter and the commonly used deep learning golden standard of using an ensemble of trained models \cite{lakshminarayanan2017simple, vyas2018out}.

\section{Method}
We start by introducing both approaches, dynamical systems based on autoencoders and their attractors and uncertainty estimation by MC dropout. Next we introduce our method, which we call Monte-Carlo Attractor Autoencoder (MCA-AE), combining both of the aforementioned design choices.

\subsection{Preliminaries - Attractors}
\label{sec:attractors}
A good overview on the basic analysis of autoencoders, associative memory and attractors is provided in \cite{radhakrishnan2020overparameterized}. Let $f$ be an autoencoder trained under the standard training regime, i.e. minimizing the reconstruction loss $\mathcal{L}$ between input $x$ and target $f(x)$, i.e. $\mathcal{L}=\mathrm{r}\left(f(x)-x\right)$, where $\mathrm{r}(\cdot)$ is a reconstruction loss of choice. Consider an input sample $x$, an index set $\mathcal{I}=\{1,2,\dots,N\}$ for some $N \geq 1$ and the sequence $\{f^k(x)\}_{k \in \mathcal{I}}$, where $f^k(x)=\left( f \circ f \circ \cdots \circ f \right)(x)$ (k times) denotes $k$ compositions of $f$ applied to $x$. A point $x$ is a \textbf{fixed point} $x^*$ of $f$ if $f(x)=x$, where we allow the equality to be weakened, i.e. $f(x) = x+\epsilon \approx x$ for some small $\epsilon$, because the reconstruction will never be perfect. The sequence $\{f^k(x)\}_{k \in \mathcal{I}}$ then converges to $x^*$. A fixed point $x^*$ is an \textbf{attractor} of $f$ if there exists an open neighborhood $\mathcal{O}$ around $x^*$ such that for all $x \in \mathcal{O}$ the sequence $\{f^k(x)\}_{k \in \mathcal{I}}$ converges to $x^*$ if $k \to \infty$. The set of all such points is called the \textbf{basin of attraction} of $x^*$ for $f$. Even disturbed training samples converge to the initial training sample \cite{radhakrishnan2020overparameterized}. We show that this property can be used to generalize to test samples, when they are close enough to the training distribution. If the latter is violated, the sample might not be stable in its convergence, which will be exploited by our next design choice.   

\subsection{Preliminaries - MC Dropout}
\label{sec:mc-d}
The use of dropout during training and inferences, called Monte Carlo (MC) dropout, has been introduced \cite{gal2016dropout} to model uncertainty in neural networks without sacrificing complexity or test accuracy for several machine learning tasks. For standard classification or regression models, an individual binary mask is sampled for each layer (except the last layer) for each new training and test sample. Consequently, neurons are dropped randomly such that during inference we sample a function $f$ from a family, or distribution of functions $\mathcal{F}$, i.e.  $f \in \mathcal{F}$. Uncertainty and reliability can then be assessed by performing multiple runs for the same input sample $x$, i.e. retrieve $\{f_j(x)\}_{j \in \mathcal{J}}$ for $\mathcal{J}=\{1,2,\cdots,M\}$ for some $M \geq 1$. The models predictive distribution for an input sample $x$ can then be assessed by computing $p=f(x) = \frac{1}{M} \sum_{j=1}^{M} \mathrm{softmax}(f_j(x))$. Uncertainty can be summarized by computing the normalized entropy \cite{laves2020calibration} of the probability vector $p$, i.e. $H(p)=-\frac{1}{\log(C)}\sum_{c=1}^{C}p_c\log(p_c)$, where $C$ is the number of classes. We use the latter in all our experiments to compute the uncertainty of the prediction and decide based on its value whether a sample is rejected or accepted for prediction or whether the sample is in- or out-of-distribution.

\subsection{MCA-AE}
\label{sec:mca-ae}
Our introduced method is a combination of both previously detailed model designs. Instead of training the autoencoder model under a standard training regime as done by related works thus far, we train the model using dropout and enabling dropout during inference as well. This causes an interesting model feature: if we repeat the recursive application of the trained autoencoder several times for the same input sample $x$, then each iteration uses a different function $f$ from the same distributions of functions $\mathcal{F}$. Hence, we obtain different, but similar, dynamical systems for inference which should behave similarly for training and test samples, but not consistently for novel feature variations in the input. Each iteration can hence converge to a different attractor, potentially of different classes. The latter is useful to detect inconsistencies and hence uncertainty: if the model converges to attractors of the same class we can assume a trustful prediction, if it converges to attractors of different classes the convergence is unreliable.

\noindent\textbf{MCA-AE:} Let $x$ be an input sample and $\mathcal{F}$ be the family of functions consisting of autoencoders learned by using dropout during training and enabling it during inference as well. We repeat the recursion $M$ times, sampling each time a new $f_j$ for each recursion $\mathcal{J}=\{1,2,\cdots,M\}$. This results in a predictive distribution $\{f_j^k(x)\}_{j \in \mathcal{J}}$, where $k$ is the number of compositions performed for each recursion. As a reminder, for a fixed $f_j$ the dropout mask is the same for each recursive step $k$. The latter implies that the dropout mask needs to be implemented manually such that it can be fixed for multiple inferences. Since we are adopting this strategy for autoencoders, we refrain from using dropout in the latent space. Classification of the resulting iteratively reconstructed sample is performed in the latent space of the $k$th iteration. For the latter we use a MLP classifier with a single hidden layer of the same size as the latent dimension. To summarize this heuristic: 

\begin{algorithmic}[1]
  \STATE Train autoencoder model using dropout to get $\mathcal{F}$
  \STATE Enable dropout for inference
  \STATE Define the number of recursions $N$
  \STATE Train classifier $\mathrm{g}(\cdot)$ in latent space after $N$ recursions
  \STATE Define the number of inferences per sample $M$
  \STATE Define uncertainty threshold $U$
  \FOR {each input sample $x$}
    \FOR{$l \gets 1$ to $M$}   
      \FOR{$k \gets 1$ to $N$}   
        \IF {$k = 1$}
          \STATE Sample a new dropout mask and keep it fixed
          \STATE This gets you $f_j \in \mathcal{F}$, where $f_j(x)=d_j(e_j(x))$
        \ENDIF
        \STATE $z = e_j(x)$ \COMMENT{encoding}
        \STATE $x = d_j(z)$ \COMMENT{decoding}
      \ENDFOR
    \STATE $y_l = g(z)$ \COMMENT{probability distribution of classification}
    \ENDFOR
    \STATE $p(y) = \frac{1}{M} \sum_{l=1}^{M} y_l$
    \STATE $H(p(y))=-\frac{1}{\log(C)}\sum_{c=1}^{C}p_c(y)\log(p_c(y))$
    \IF {$H(p(y)) \leq U$}
      \STATE 
        $y = \mathrm{argmax}(y_l)$
      \ELSE
      \STATE Reject sample
    \ENDIF
  \ENDFOR
\end{algorithmic}

For training samples to become attractors it is necessary to train the autoencoders for a large number of epochs, i.e. we used 25000. A lot of hyperparameters are defined for the inference process instead of the training process. The number of recursions and the number of different runs is independent from the training. The classifier can be chosen after the autoencoder training. The uncertainty threshold needs to be adapted according to the use case and it is a tradeoff between the required sensitivity and precision. 

\section{Experiments}
We evaluate our method on two scenarios: First, we want to assess the predictive uncertainty where the model should provide a high uncertainty in case it wrongly classifies a sample. This is made more difficult in the case of the vehicle interior: unseen objects should be classified as empty seats, i.e. the model should only identify known classes and neglect everything else. Our results will show that this is a challenging task. Second, the model should differentiate between in- and out-of-distribution (OOD) samples. In the case of training on MNIST and evaluating on Fashion-MNIST, the model cannot perform a correct prediction and it should detect the OOD as such. This is also the case when images from a new vehicle interior are provided as input to the model. All training and evaluation scripts can be found in \href{https://github.com/SteveCruz/icpr2022-autoencoder-attractors}{our implementation (link)}.

\subsection{Evaluation metrics}
\label{sec:metrics}
According to standard evaluation criterions adopted in related works, we evaluate our models using the Area Under the Receiver Operating Characteristic curve (AUROC), Area Under the Precision-Recall curve (AUPR) and the false positive rate at 95\% true positive rate (FPR95\%). For OOD evaluation we use approximately 50\% of the samples from the test set $\mathcal{D}_{in}$ and 50\% from the test set from $\mathcal{D}_{out}$. For further details and interpretations of the metrics we refer to \cite{hendrycks17baseline, hendrycks2018deep, davis2006relationship, manning1999foundations, liu2018open}. 

\subsection{Datasets}
We use several commonly used computer vision datasets for training and use the corresponding test data as in-distribution sets $\mathcal{D}_{in}$: MNIST \cite{lecun1998gradient}, Fashion-MNIST \cite{xiao2017fashion}, SVHN \cite{37648} and GTSRB \cite{Houben-IJCNN-2013} (which we reduce to use 10 classes only). For out-of-distribution $\mathcal{D}_{out}$ we use a subset of all $\mathcal{D}_{in}$ not coming from the training distribution and the test datasets from Omniglot \cite{lake2015human}, CIFAR10 \cite{krizhevsky2009learning}, LSUN \cite{yu2015lsun} (for which we use the train split) and Places365 \cite{zhou2017places}. We use approximately the same number of samples from $\mathcal{D}_{in}$ and $\mathcal{D}_{out}$ by sampling each class uniformly. An overview is provided in Table \ref{table:datasets}. 
\begin{table}
  \caption{Overview of the number of classes and samples for OOD or uncertainty estimation for the different datasets used.}
  \label{table:datasets}
  \centering
  \begin{tabular}{lrrr}
    \toprule
    Dataset & Classes & $D_{in}$ and $D_{out}$ & Uncertainty \\
    \midrule
    MNIST & 10 & 2500 & 10000 \\
    Fashion & 10 & 2500 & 26032 \\
    SVHN & 10 & 2500 & 10000 \\
    GTSRB & 10 & 2006 & 3208 \\
    CIFAR10 & 10 & 2500 & - \\
    Omniglot & 660 & 2636 & -\\
    LSUN & 10 & 2500 & -\\
    Places365 & 365 & 2555 & - \\
    SVIRO-U Adults (A) & 7 & 1337 & 2617 \\
    SVIRO-U Seats (S) & 8 & - & 490 \\
    SVIRO-U Objects (O) & 8 & - & 1622 \\
    SVIRO-U A,S & 26 & - & 896 \\
    SVIRO-U A,O & 7 & - & 1421 \\
    SVIRO-U A,S,O & 30 & - & 1676 \\
    SVIRO Tesla & 21 & - & 2000 \\
    \bottomrule
  \end{tabular}
\end{table}

\begin{figure}
  \centering
  \includegraphics[width=2.1cm]{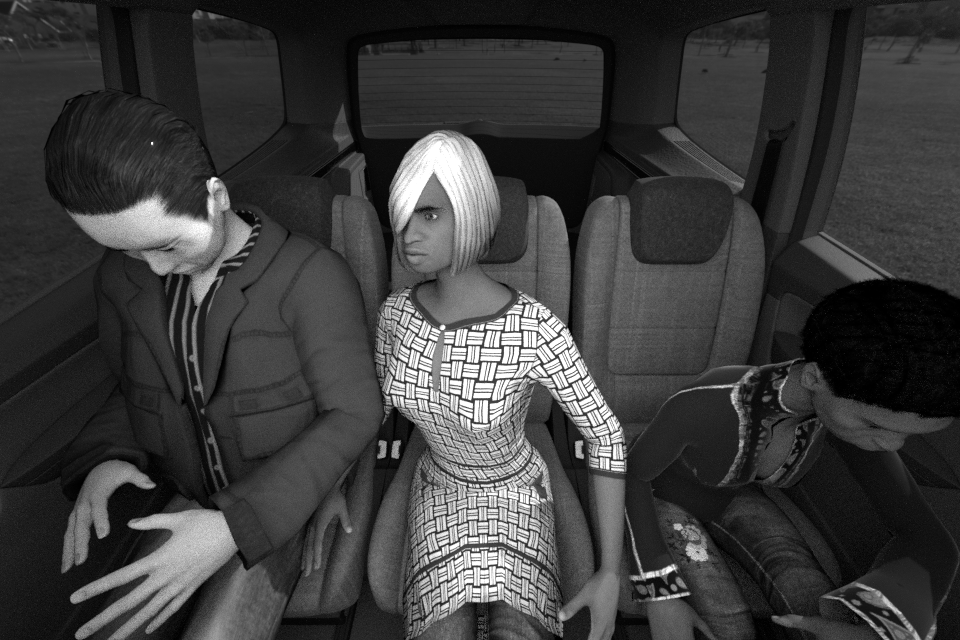}
  \includegraphics[width=2.1cm]{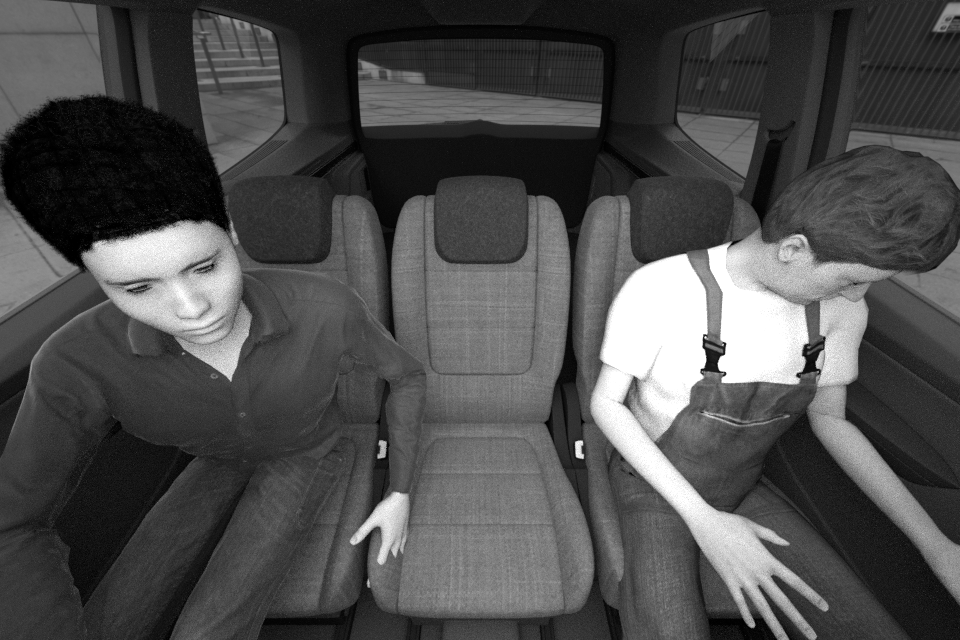}
  \includegraphics[width=2.1cm]{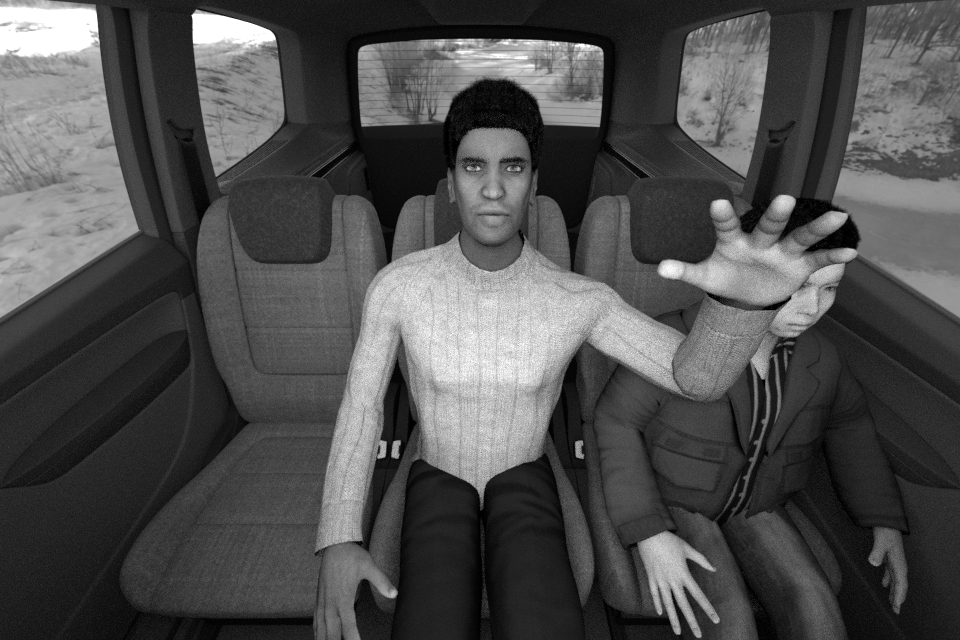}
  \includegraphics[width=2.1cm]{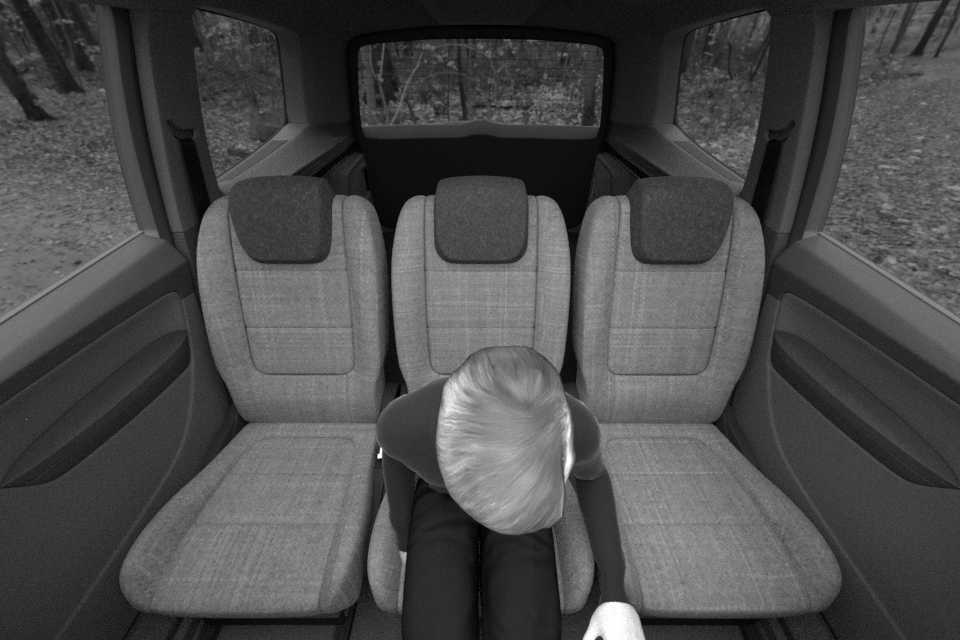}\\
  \vspace{0.1cm}
  \includegraphics[width=2.1cm]{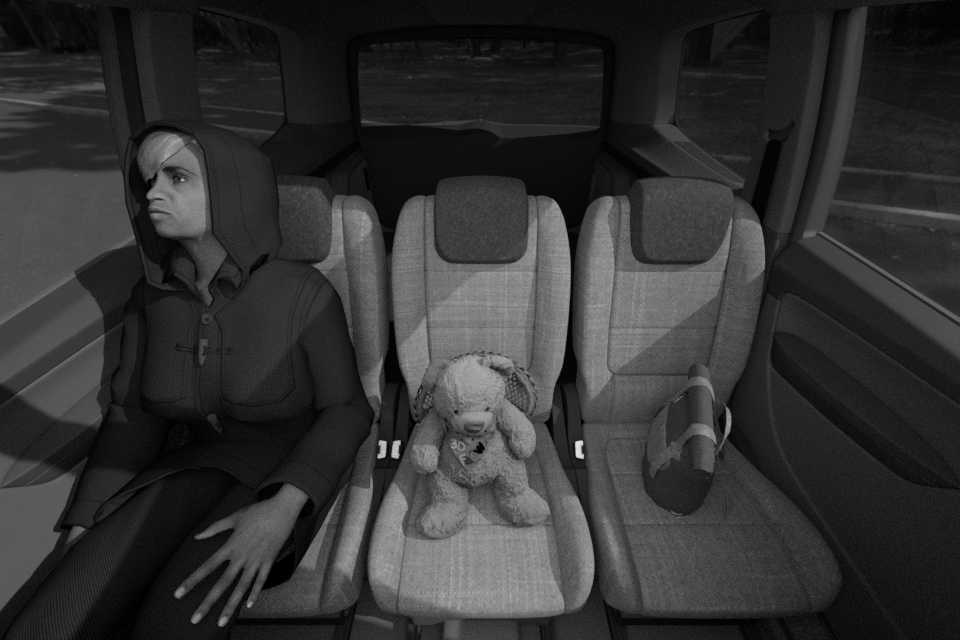}
  \includegraphics[width=2.1cm]{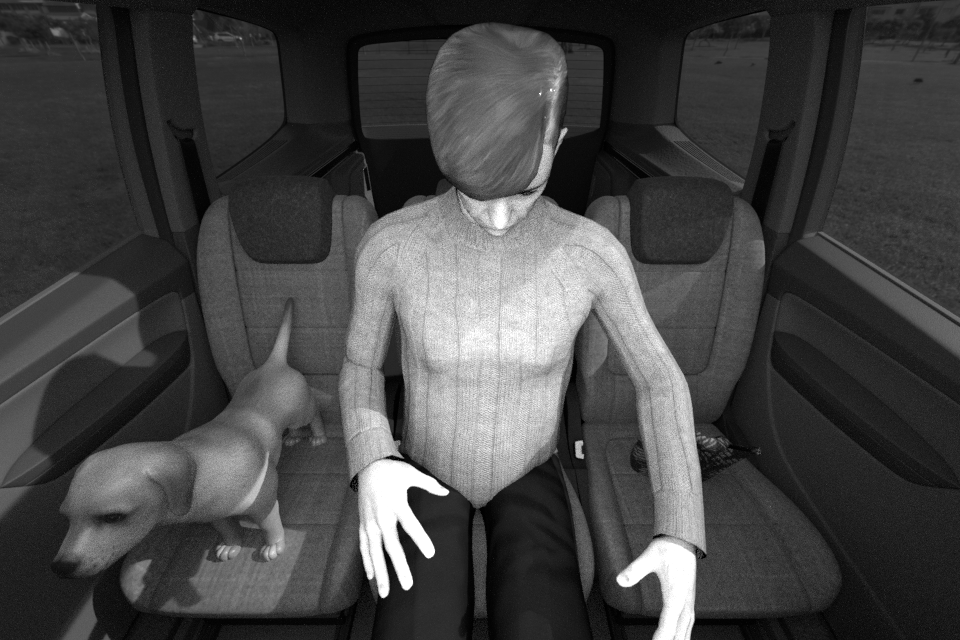}
  \includegraphics[width=2.1cm]{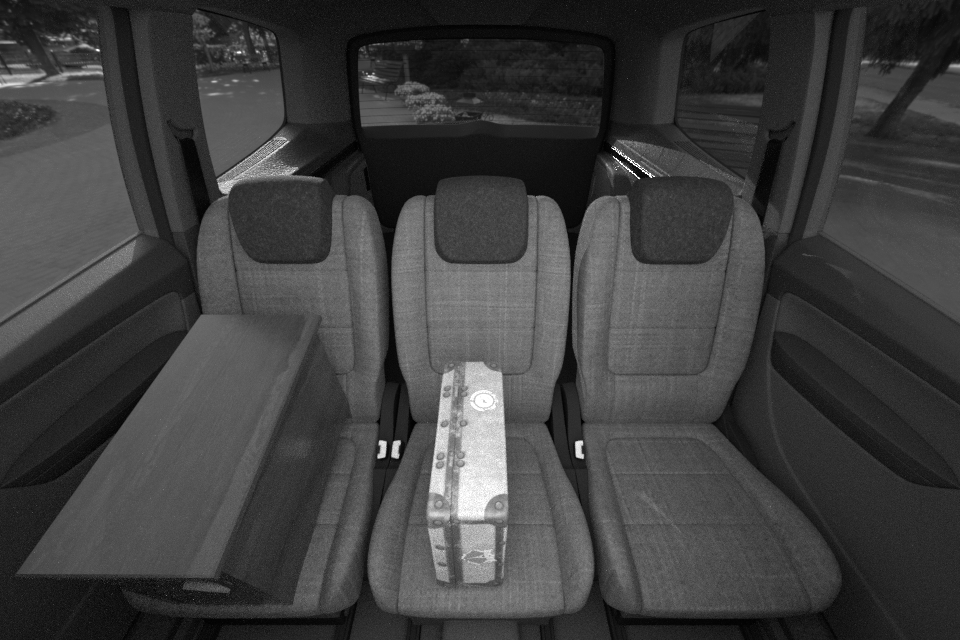}
  \includegraphics[width=2.1cm]{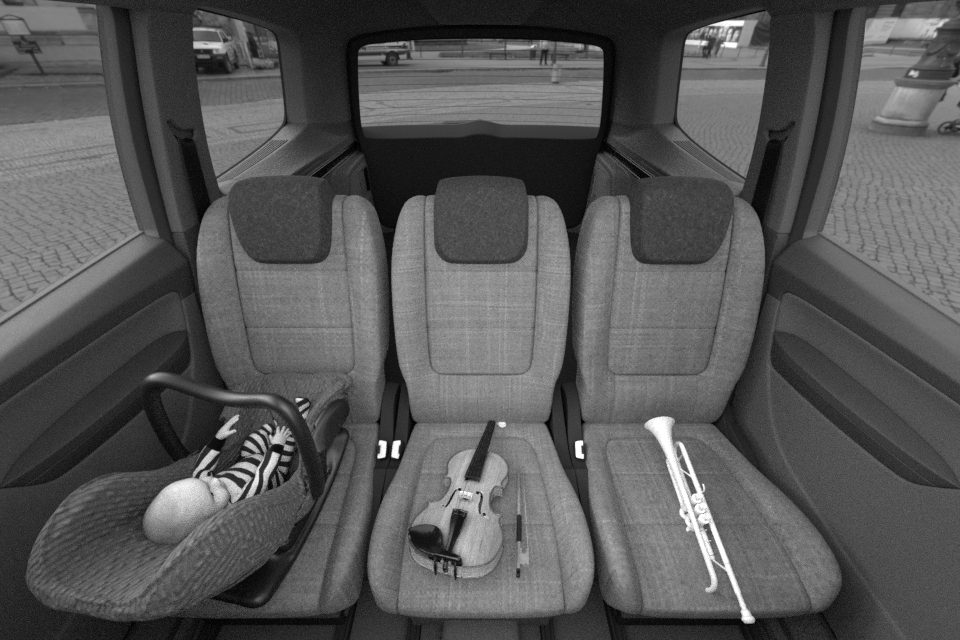}
  \caption{Examples from the SVIRO-Uncertainty dataset. First row are training samples of adults only. Second row are test samples of unseen adults, but also child-seats and everyday objects which should be classified as empty.}
  \label{fig:sviro}
\end{figure}

In addition to these commonly used datasets, we release an extension for SVIRO \cite{DiasDaCruz2020SVIRO} called SVIRO-Uncertainty. For each of the 3 seat positions in the vehicle interior rear bench the model should classify which object is occupying it, with empty being one possible choice. We created two training datasets for the Sharan vehicle, one using adult passengers only (4384 sceneries and 8 classes) and one using adults, child seats and infant seats (3515 samples and 64 classes - not used for training in this work). We created fine-grained test sets to asses the reliability on several difficulty levels: 1) only unseen adults, 2) only unseen child and infant seats, 3) unseen adults and unseen child and infant seats, 4) unknown random everyday objects (e.g. dog, plants, bags, washing machine, instruments, tv, skateboard, paintings, ...), 5) unseen adults and unknown everyday objects and 6) unseen adults, unseen child and infant seats and unknown everyday objects. The dataset \href{https://sviro.kl.dfki.de/sviro-uncertainty-download/}{can been downloaded (link)}. Besides the uncertainty estimation within the same vehicle interior, one can use images from unseen vehicle interiors from SVIRO to further test the models reliability on the same task, but in novel environments, i.e. vehicle interiors. Example images are provided in Fig. \ref{fig:sviro}.

\begin{table*}
  \caption{Comparison (in percentage) of our method against MC dropout and an ensemble of models. We repeated the experiments for 10 runs and report the mean values together with their standard deviation. If $\mathcal{D}_{in}=\mathcal{D}_{out}$, then we report the result on the test set of $\mathcal{D}_{in}$ only. Arrows indicate whether larger $\uparrow$ or smaller $\downarrow$ is better. Best results are highlighted in grey. The last block is a comparison on the fine-grained splits on the newly released SVIRO-Uncertainty. All but adults should be classified as empty.}
  \label{table:ood}
  \centering
  \begin{tabular}{l@{\quad}S[table-format=2.1(1)]@{\quad}S[table-format=2.1(1)]@{\quad}S[table-format=2.1(1)]@{\quad}S[table-format=2.1(1)]@{\quad}S[table-format=2.1(1)]@{\quad}S[table-format=2.1(1)]@{\quad}S[table-format=2.1(1)]@{\quad}S[table-format=2.1(1)]@{\quad}S[table-format=2.1(1)]}
    \toprule
    & \multicolumn{3}{c}{MCA-AE (Ours)} & \multicolumn{3}{c}{MC Dropout} & \multicolumn{3}{c}{Ensemble of 10 models} \\
    \midrule
    $\mathcal{D}_{in} \to \mathcal{D}_{out}$ & AUROC \hspace{3pt} $\uparrow$ & AUPR \hspace{3pt} $\uparrow$ & FPR\SI{95}{\percent} \hspace{3pt} $\downarrow$ & AUROC \hspace{3pt} $\uparrow$ & AUPR \hspace{3pt} $\uparrow$ & FPR\SI{95}{\percent} \hspace{3pt} $\downarrow$ & AUROC \hspace{3pt} $\uparrow$ & AUPR \hspace{3pt} $\uparrow$ & FPR\SI{95}{\percent} \hspace{3pt} $\downarrow$ \\
    \midrule
     MNIST &&&&&&&&& \\
     $\to$MNIST    & 79.9 \pm 1.6 & 94.9 \pm 0.5 & 74.2 \pm 4.8 &        $\cellcolor{good}$ 90.1 \pm 0.6 & $\cellcolor{good}$ 99.6 \pm 0.1 & $\cellcolor{good}$ 28.0 \pm 2.4 &                       85.8 \pm 1.4 & 99.0 \pm 0.1 & 41.5 \pm 3.0 \\
     $\to$CIFAR10  & 88.2 \pm 2.3 & 87.4 \pm 2.2 & 44.1 \pm 8.3 &        91.2 \pm 1.3 & 92.2 \pm 1.1 & 39.8 \pm 5.1 &                                                           $\cellcolor{good}$ 91.5 \pm 1.1 & $\cellcolor{good}$ 92.4 \pm 0.9 & $\cellcolor{good}$ 34.0 \pm 4.7 \\
     $\to$Fashion  & 74.5 \pm 3.2 & 72.7 \pm 3.3 & 72.3 \pm 4.4 &                         $\cellcolor{good}$ 90.0 \pm 1.6 & $\cellcolor{good}$ 91.1 \pm 1.3 &  40.5 \pm 5.9 &         89.5 \pm 1.1 & 90.6 \pm 0.9 & $\cellcolor{good}$ 37.0 \pm 3.2 \\
     $\to$Omniglot & 64.4 \pm 5.0 & 70.4 \pm 5.7 & 99.4 \pm 0.7 &                          93.4 \pm 2.8 & 94.2 \pm 2.5 & 35.4 \pm 12.2 &                                                            $\cellcolor{good}$ 95.5 \pm 1.0 & $\cellcolor{good}$ 96.0 \pm 0.8 & $\cellcolor{good}$ 22.0 \pm 6.0 \\
     $\to$SVHN     & 92.2 \pm 2.0 & 91.3 \pm 1.5 & 28.2 \pm 12.4 &        94.2 \pm 1.7 & 94.9 \pm 1.4 & 30.5 \pm 9.4 &                                                            $\cellcolor{good}$ 94.9 \pm 0.9 & $\cellcolor{good}$ 95.4 \pm 0.7 & $\cellcolor{good}$  22.4 \pm 5.1 \\
     \midrule
     Fashion &&&&&&&&& \\
     $\to$Fashion  & 81.0 \pm 1.0 & 94.4 \pm 0.3 & 80.2 \pm 2.4 &                                          $\cellcolor{good}$ 82.5 \pm 0.4 & $\cellcolor{good}$ 96.4 \pm 0.1 & $\cellcolor{good}$ 64.4 \pm 4.3 &          81.7 \pm 0.7 & $\cellcolor{good}$ 96.4 \pm 0.1 & 64.7 \pm 2.2 \\
     $\to$CIFAR10  & $\cellcolor{good}$ 93.9 \pm 1.8 & $\cellcolor{good}$ 95.8 \pm 1.1 & 47.0 \pm 20.0 &        88.7 \pm 1.9 & 89.6 \pm 1.8 & 45.7 \pm 5.8 &                                            91.6 \pm 0.9 & 92.1 \pm 0.8 & $\cellcolor{good}$ 34.3 \pm 3.1 \\
     $\to$MNIST    & 87.8 \pm 4.0 & 88.2 \pm 3.4 & 48.7 \pm 15.6 &                                          85.4 \pm 1.8 & 86.7 \pm 1.5 & 53.5 \pm 5.1 &                                            $\cellcolor{good}$ 90.2 \pm 0.5 & $\cellcolor{good}$ 90.6 \pm 0.5 & $\cellcolor{good}$ 35.7 \pm 2.4 \\
     $\to$Omniglot & 86.8 \pm 3.8 & 91.2 \pm 2.5 & 87.3 \pm 9.7 &                                          93.6 \pm 2.0 & 94.1 \pm 1.8 & 32.6 \pm 10.1 &                                           $\cellcolor{good}$ 97.9 \pm 0.4 & $\cellcolor{good}$ 98.1 \pm 0.3 & $\cellcolor{good}$ 9.3 \pm 2.3 \\
     $\to$SVHN     & 93.7 \pm 2.0 & $\cellcolor{good}$ 95.6 \pm 1.3 & 48.9 \pm 17.1 &                       90.8 \pm 1.0 & 91.7 \pm 0.9 & 40.7 \pm 3.6 &                        $\cellcolor{good}$  94.8 \pm 0.5 & 95.1 \pm 0.4 & $\cellcolor{good}$ 23.0 \pm 2.6 \\
     \midrule
     SVHN &&&&&&&&& \\
     $\to$SVHN      & 77.6 \pm 0.8 & 80.8 \pm 1.0 & 79.5 \pm 2.1 &                        $\cellcolor{good}$ 84.0 \pm 0.6 & $\cellcolor{good}$ 93.1 \pm 0.4 & 69.3 \pm 2.4 &          83.7 \pm 0.5 & 92.9 \pm 0.3 & $\cellcolor{good}$ 68.7 \pm 2.0 \\
     $\to$CIFAR10   & 77.5 \pm 1.2 & 80.4 \pm 1.1 &  83.6 \pm 3.0 &                       74.9 \pm 0.9 & 78.0 \pm 0.8 & 85.8 \pm 1.8 &                                                          $\cellcolor{good}$ 77.6 \pm 0.7 & $\cellcolor{good}$ 80.5 \pm 0.6 & $\cellcolor{good}$ 83.3 \pm 1.4 \\
     $\to$GTSRB     & $\cellcolor{good}$ 75.4 \pm 2.2 & 80.5 \pm 1.9 & $\cellcolor{good}$ 80.7 \pm 5.4 &      74.0 \pm 1.1 & 80.1 \pm 1.0 & 84.9 \pm 2.9 &                                                           75.3 \pm 0.7 & $\cellcolor{good}$ 81.2 \pm 0.7 & 84.0 \pm 3.0 \\
     $\to$LSUN      & 78.4 \pm 0.9 & 81.5 \pm 0.8 & 82.7 \pm 4.8 &                       77.0 \pm 0.7 & 79.8 \pm 0.7 & 81.9 \pm 2.1 &                                                           $\cellcolor{good}$ 79.2 \pm 0.7 & $\cellcolor{good}$ 81.9 \pm 0.7 & $\cellcolor{good}$ 80.1 \pm 1.9 \\
     $\to$Places365 & 78.5 \pm 0.8 & 81.0 \pm 0.7 & 82.6 \pm 3.7 &                      77.1 \pm 0.6 & 79.4 \pm 0.6 & 80.9 \pm 2.5 &                                                            $\cellcolor{good}$ 79.2 \pm 0.5 & $\cellcolor{good}$ 81.5 \pm 0.4 & $\cellcolor{good}$ 79.5 \pm 1.9 \\
     \midrule
     GTSRB &&&&&&&&& \\
     $\to$GTSRB     & 85.1 \pm 0.9 & 95.6 \pm 0.5 &  69.3 \pm 3.3 &       $\cellcolor{good}$ 89.3 \pm 2.4 & $\cellcolor{good}$ 98.8 \pm 0.3 &  $\cellcolor{good}$ 50.9 \pm 6.0 &          84.6 \pm 1.7 & 97.4 \pm 0.3 & 62.1 \pm 3.2 \\
     $\to$CIFAR10   & $\cellcolor{good}$ 91.4 \pm 0.6 & $\cellcolor{good}$ 90.3 \pm 0.8 & $\cellcolor{good}$ 42.0 \pm 3.3 &       81.2 \pm 0.9 & 81.4 \pm 0.9 & 69.5 \pm 3.7 &          76.3 \pm 0.5 & 77.7 \pm 0.5 & 83.4 \pm 1.3 \\
     $\to$LSUN      & $\cellcolor{good}$93.0 \pm 0.7 & $\cellcolor{good}$ 92.2 \pm 0.7 & $\cellcolor{good}$ 36.5 \pm 4.4 &       83.4 \pm 0.8 & 83.3 \pm 0.7 & 65.3 \pm 3.9 &          77.7 \pm 0.8 & 78.7 \pm 0.6 & 81.3 \pm 1.6 \\
     $\to$Places365 & $\cellcolor{good}$ 92.3 \pm 0.7 & $\cellcolor{good}$ 91.3 \pm 0.7 & $\cellcolor{good}$ 38.8 \pm 3.4 &       82.8 \pm 0.7 & 82.4 \pm 0.6 & 65.1 \pm 3.6 &          77.5 \pm 0.6 & 78.2 \pm 0.6 & 80.6 \pm 1.7 \\
     $\to$SVHN      & $\cellcolor{good}$ 91.3 \pm 0.7 &$\cellcolor{good}$ 90.7 \pm 0.8 & $\cellcolor{good}$ 44.5 \pm 3.7 &       85.6 \pm 1.5 & 85.5 \pm 1.5 & 60.7 \pm 5.1 &          79.4 \pm 0.6 & 80.3 \pm 0.5 & 80.1 \pm 1.7 \\
     \midrule
     SVIRO-U &&&&&&&&& \\
     $\to$CIFAR10   & $\cellcolor{good}$ 95.4 \pm 0.6 & $\cellcolor{good}$ 93.3 \pm 1.0 & $\cellcolor{good}$ 26.9 \pm 3.4 &       74.6 \pm 3.5 & 73.6 \pm 2.0 & 60.4 \pm 7.2 &          77.7 \pm 1.5 & 75.0 \pm 1.1 & 57.1 \pm 3.2 \\
     $\to$GTSRB     & $\cellcolor{good}$ 95.8 \pm 1.0 & $\cellcolor{good}$ 94.9 \pm 1.1 & $\cellcolor{good}$ 25.1 \pm 6.9 &       69.9 \pm 2.7 & 74.2 \pm 1.2 & 68.8 \pm 4.7 &          74.7 \pm 2.5 & 76.2 \pm 1.5 & 63.8 \pm 1.3 \\
     $\to$LSUN      & $\cellcolor{good}$ 94.8 \pm 0.5 & $\cellcolor{good}$ 92.7 \pm 0.7 & $\cellcolor{good}$ 31.5 \pm 2.7 &       67.6 \pm 2.0 & 70.1 \pm 1.0 & 72.3 \pm 4.2 &          72.0 \pm 1.1 & 71.6 \pm 0.6 & 64.4 \pm 2.3 \\
     $\to$Places365 & $\cellcolor{good}$ 95.4 \pm 0.5 & $\cellcolor{good}$ 93.3 \pm 0.7 & $\cellcolor{good}$ 27.3 \pm 2.8 &       73.2 \pm 2.6 & 72.5 \pm 1.3 & 63.5 \pm 6.8 &          77.4 \pm 1.0 & 74.5 \pm 0.7 & 57.0 \pm 2.5 \\
     $\to$SVHN      & $\cellcolor{good}$ 92.4 \pm 1.6 & $\cellcolor{good}$ 88.6 \pm 2.3 & $\cellcolor{good}$ 40.1 \pm 7.6 &       81.0 \pm 3.4 & 77.8 \pm 2.4 & 49.5 \pm 8.9 &          81.0 \pm 1.3 & 77.3 \pm 1.1 & 51.6 \pm 4.1 \\
    \midrule
     $\to$Adults (A)   & 87.8 \pm 1.3 & 99.1 \pm 0.3 &  62.9 \pm 3.9 &     $\cellcolor{good}$ 95.2 \pm 1.7 &        $\cellcolor{good}$ 99.9 \pm 0.1 & $\cellcolor{good}$ 8.9 \pm 3.1 &                            91.1 \pm 1.9 &  99.8 \pm 0.1 & 28.8 \pm 8.8 \\
     $\to$Seats (S)    & $\cellcolor{good}$ 54.0 \pm 7.5 & $\cellcolor{good}$ 8.9 \pm 4.2 & $\cellcolor{good}$ 88.8 \pm 10.8     &            17.5 \pm 13.1 & 0.4 \pm 0.2 & 95.7 \pm 5.1 &         28.1 \pm 6.8 & 2.6 \pm 1.6 & 98.0 \pm 2.5 \\
     $\to$Objects (O)  & $\cellcolor{good}$ 68.9 \pm 3.1 & 83.7 \pm 2.4 & $\cellcolor{good}$ 84.1 \pm 5.5 &    64.7 \pm 3.2 &         $\cellcolor{good}$ 85.3 \pm 3.3 & 85.3 \pm 4.6 &          57.4 \pm 2.3 & 80.6 \pm 1.4 & 86.5 \pm 3.3 \\
     $\to$A,S          & $\cellcolor{good}$ 58.8 \pm 2.6 & $\cellcolor{good}$ 48.6 \pm 6.4 & $\cellcolor{good}$ 93.2 \pm 1.1 &          36.3 \pm 2.3 & 16.5 \pm 2.8 & 97.4 \pm 1.4 &          39.5 \pm 1.7 & 23.6 \pm 1.8 & 96.9 \pm 1.1 \\
     $\to$A,O          & $\cellcolor{good}$ 78.8 \pm 1.5 & 93.0 \pm 0.5 & $\cellcolor{good}$ 76.1 \pm 3.4 &    70.1 \pm 1.6 &         $\cellcolor{good}$ 93.5 \pm 0.9 & 77.4 \pm 2.8 &          71.1 \pm 0.7 & 92.3 \pm 0.6 & 77.8 \pm 2.4 \\
     $\to$A,S,O        & $\cellcolor{good}$ 62.2 \pm 2.0 & $\cellcolor{good}$ 56.4 \pm 4.7 & $\cellcolor{good}$ 88.7 \pm 3.1 &            42.1 \pm 2.7 & 18.6 \pm 2.7 & 96.4 \pm 0.9 &          45.8 \pm 1.9 & 33.0 \pm 1.7 & 95.5 \pm 0.9 \\
     $\to$Tesla (OOD)  & $\cellcolor{good}$ 88.6 \pm 2.0 & $\cellcolor{good}$ 97.4 \pm 0.5 &  58.0 \pm 6.1 &            52.1 \pm 2.9 & 90.9 \pm 0.5 & 94.1 \pm 3.7 &          28.6 \pm 28.6 & 45.8 \pm 45.8 & $\cellcolor{good}$ 44.4 \pm 44.4 \\
    \bottomrule
  \end{tabular}
\end{table*}

\subsection{Training and evaluation details}
We compare our method against MC dropout and an ensemble of models using the same architecture as the autoencoder encoder part, but with an additional classification head. We trained our MCA-AE models for 25000 epochs, but fewer epochs might produce good results as well. We did not perform an ablation study with respect to the number of epochs needed. Further, we did not check whether the training samples are truly fixed point and attractors because of the computational overhead: This could be done by computing the largest eigenvalue of the Jacobian matrix for each training sample and checking whether its greater than 1. The autoencoder model was trained as a denoiser \cite{xie2012image} (blur, random noise, brightness and contrast augmentation were used) to facilitate and robustify the recursive autoencoder application. Consequently, to have a fair benchmark, MC dropout and ensemble models used the same augmented images during training. The latter were trained for 1000 epochs. All methods used Adam, a learning rate of $1e^{-4}$ and a batch size of 64. For training on MNIST and Fashion-MNIST we used a latent space of 10, while for all others we used a latent space of 64. We used SSIM \cite{bergmann2018improving} for computing the reconstruction loss. We used 250 samples per class for training and treat all datasets as grayscale images. All images were centre cropped and resized to 64 pixels. We used a dropout rate of 0.33 for all methods. Model and training details can be found in the implementation.

\begin{figure*}
  \centering
  \subfloat[$D_{in}$: Fashion-MNIST]{%
    \begin{tabular}{c}%
      \begin{overpic}[height=2.4cm]{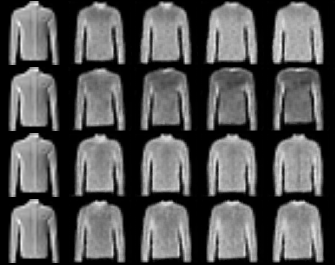}
        \put(4,73){\tiny{\textcolor{white}{Input}}}
        \put(24,73){\tiny{\textcolor{white}{Iter 1}}}
        \put(44,73){\tiny{\textcolor{white}{Iter 2}}}
        \put(64,73){\tiny{\textcolor{white}{Iter 3}}}
        \put(84,73){\tiny{\textcolor{white}{Iter 4}}}
      \end{overpic} \\[-4pt]
      \includegraphics[height=2.409cm]{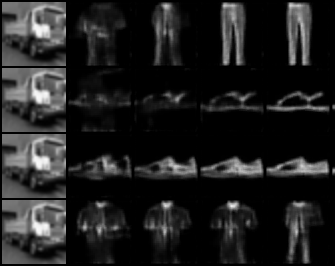}
    \end{tabular}
  }%
  \hfill
  \subfloat[$D_{in}$: MNIST]{%
    \begin{tabular}{c}%
      \begin{overpic}[height=2.409cm]{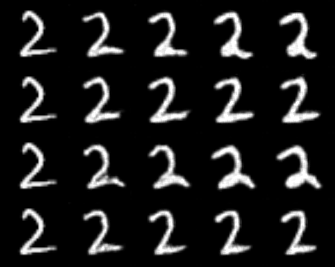}
        \put(4,73){\tiny{\textcolor{white}{Input}}}
        \put(24,73){\tiny{\textcolor{white}{Iter 1}}}
        \put(44,73){\tiny{\textcolor{white}{Iter 2}}}
        \put(64,73){\tiny{\textcolor{white}{Iter 3}}}
        \put(84,73){\tiny{\textcolor{white}{Iter 4}}}
      \end{overpic} \\[-4pt]
      \includegraphics[height=2.4cm]{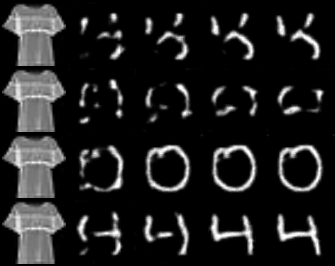}
    \end{tabular}
  }%
  \hfill
  \subfloat[$D_{in}$: SVHN]{%
    \begin{tabular}{c}%
      \begin{overpic}[height=2.4cm]{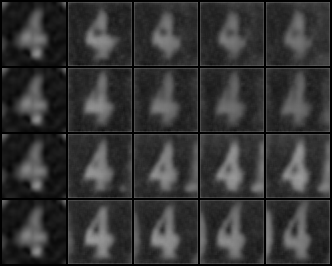}
        \put(4,73){\tiny{\textcolor{white}{Input}}}
        \put(24,73){\tiny{\textcolor{white}{Iter 1}}}
        \put(44,73){\tiny{\textcolor{white}{Iter 2}}}
        \put(64,73){\tiny{\textcolor{white}{Iter 3}}}
        \put(84,73){\tiny{\textcolor{white}{Iter 4}}}
      \end{overpic} \\[-4pt]
      \includegraphics[height=2.4cm]{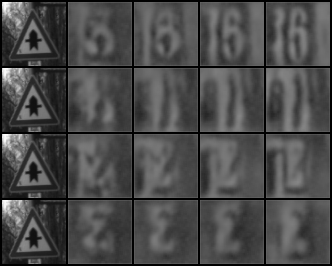}
    \end{tabular}
  }%
  \hfill
  \subfloat[$D_{in}$: GTSRB]{%
    \begin{tabular}{c}%
      \begin{overpic}[height=2.4cm]{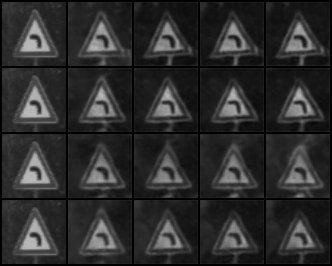}
        \put(4,73){\tiny{\textcolor{white}{Input}}}
        \put(24,73){\tiny{\textcolor{white}{Iter 1}}}
        \put(44,73){\tiny{\textcolor{white}{Iter 2}}}
        \put(64,73){\tiny{\textcolor{white}{Iter 3}}}
        \put(84,73){\tiny{\textcolor{white}{Iter 4}}}
      \end{overpic} \\[-4pt]
      \includegraphics[height=2.4cm]{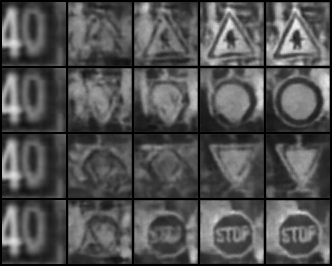}
    \end{tabular}
  }
  \hfill
  \subfloat[$D_{in}$: SVIRO-U]{%
    \begin{tabular}{c}%
      \begin{overpic}[height=2.4cm]{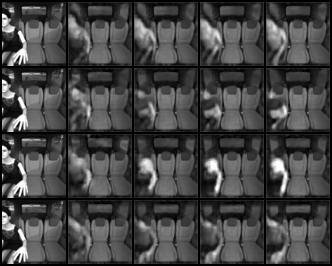}
        \put(4,73){\tiny{\textcolor{white}{Input}}}
        \put(24,73){\tiny{\textcolor{white}{Iter 1}}}
        \put(44,73){\tiny{\textcolor{white}{Iter 2}}}
        \put(64,73){\tiny{\textcolor{white}{Iter 3}}}
        \put(84,73){\tiny{\textcolor{white}{Iter 4}}}
      \end{overpic} \\[-4pt]
      \includegraphics[height=2.4cm]{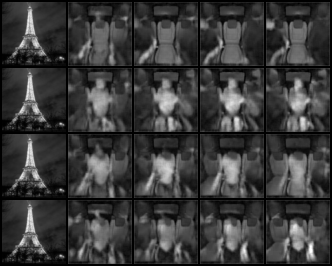}
    \end{tabular}
  }
  \caption{Multiple recursive reconstructions (from left to right) of identical samples (first column) from $D_{in}$ and $D_{out}$ by our novel MCA-AE model. Notice the evolution in the reconstruction results over each iterative step for the OOD samples. $D_{in}$ converge more robustly compared to $D_{out}$ reconstructions.}
  \label{fig:reconstructions}
\end{figure*}

For MCA-AE and MC dropout we used $20$ inferences and we used an ensemble of $10$ models to assess uncertainty and the OOD estimation. We repeated each training for 10 runs for MCA-AE and MC Dropout and for 100 runs to get the ensembles of models. We used $2$ recursions for MCA-AE, but this value depends on the dataset used and it is subject to a hyperparameter search. In our case, the models converged fast for test and slow for OOD samples, see Fig. \ref{fig:reconstructions}. Hence, more iterations did not provide an improvement. 

\subsection{Uncertainty estimation and out-of-distribution detection}

\begin{figure*}
  \centering
  \subfloat[MCA-AE (Ours)]{%
      \includegraphics[height=3.6cm]{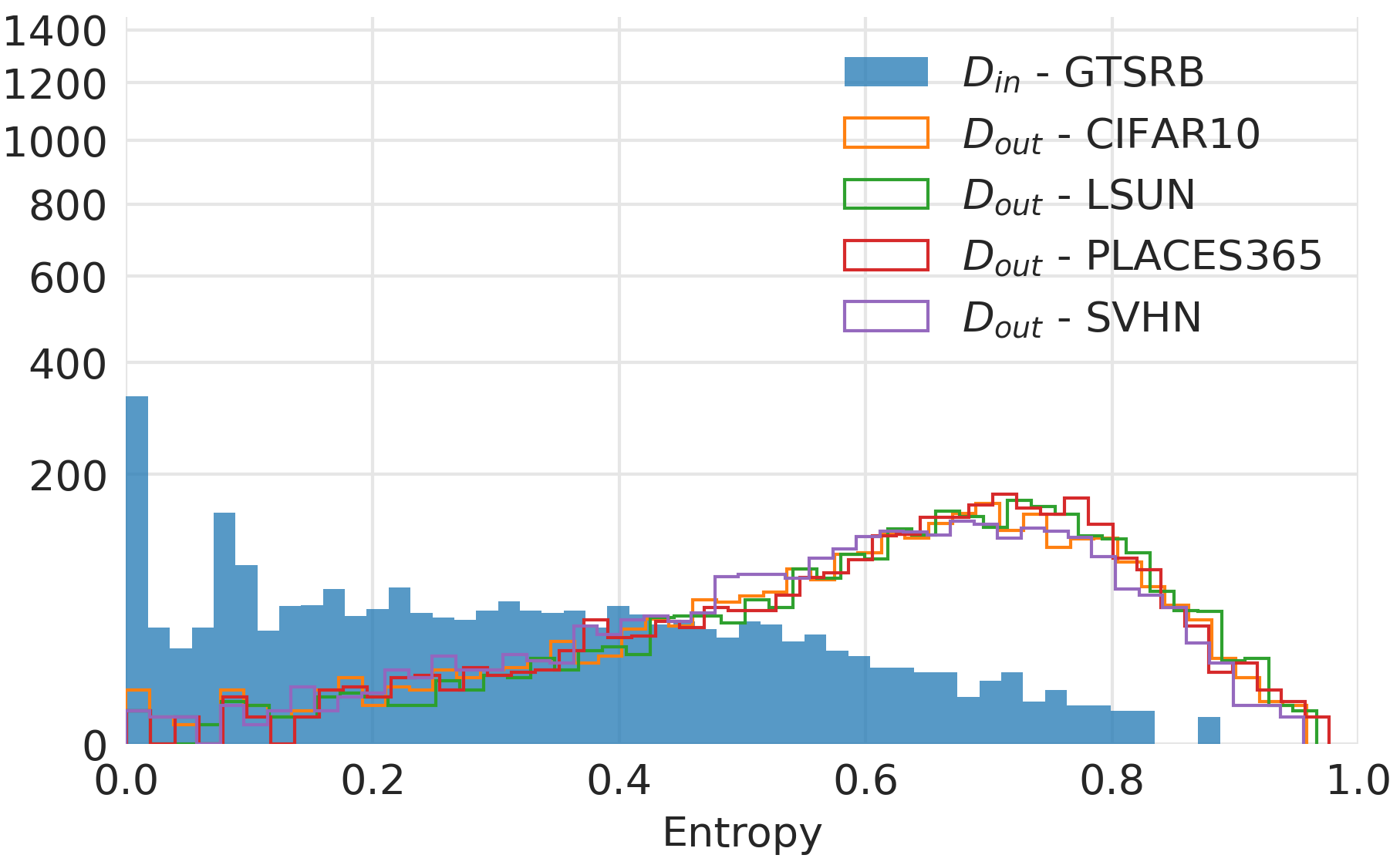}
  }%
  \hfill
  \subfloat[MC Dropout]{%
  \includegraphics[height=3.6cm]{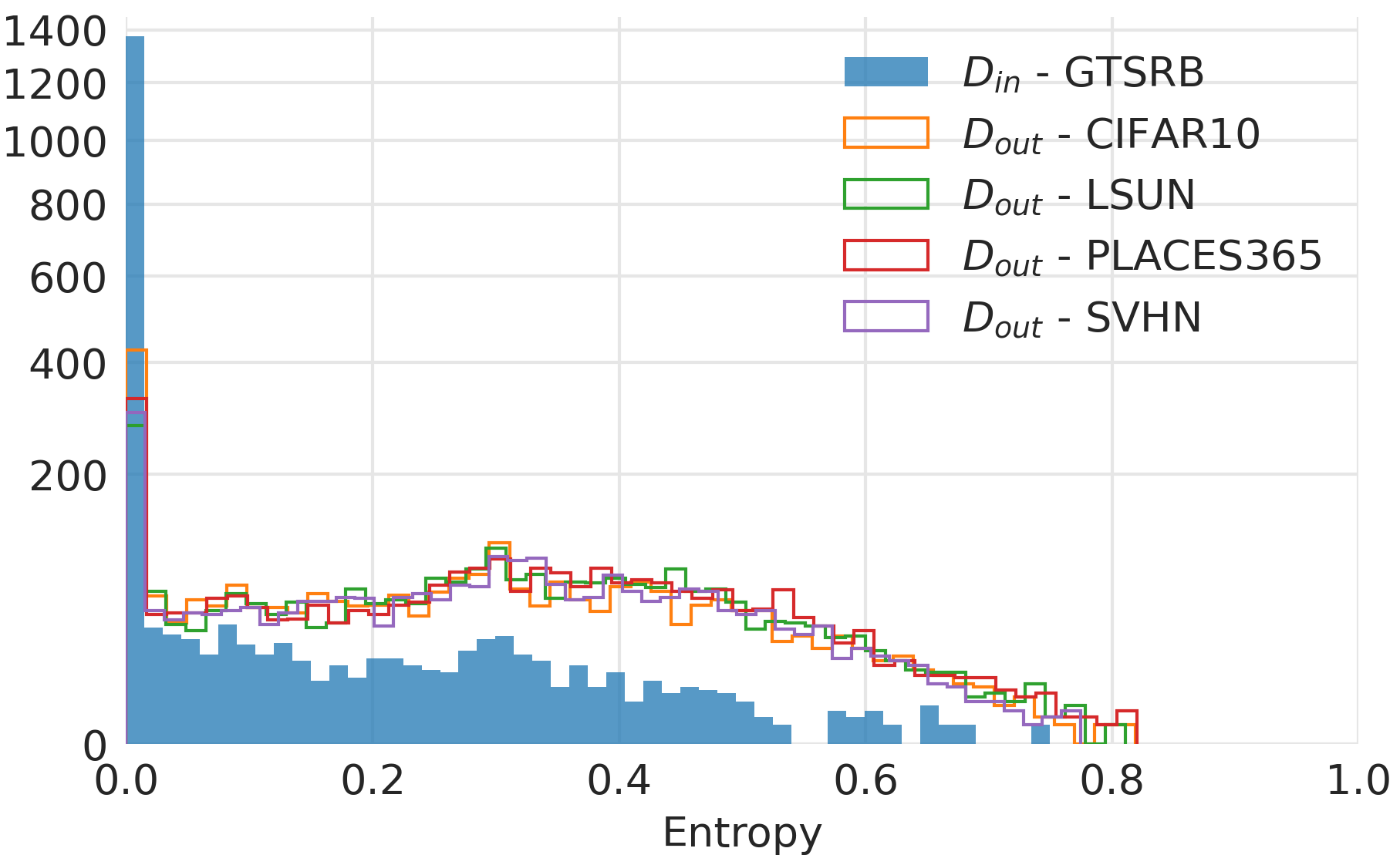}
  }%
  \hfill
  \subfloat[Ensemble of 10 models]{%
      \includegraphics[height=3.6cm]{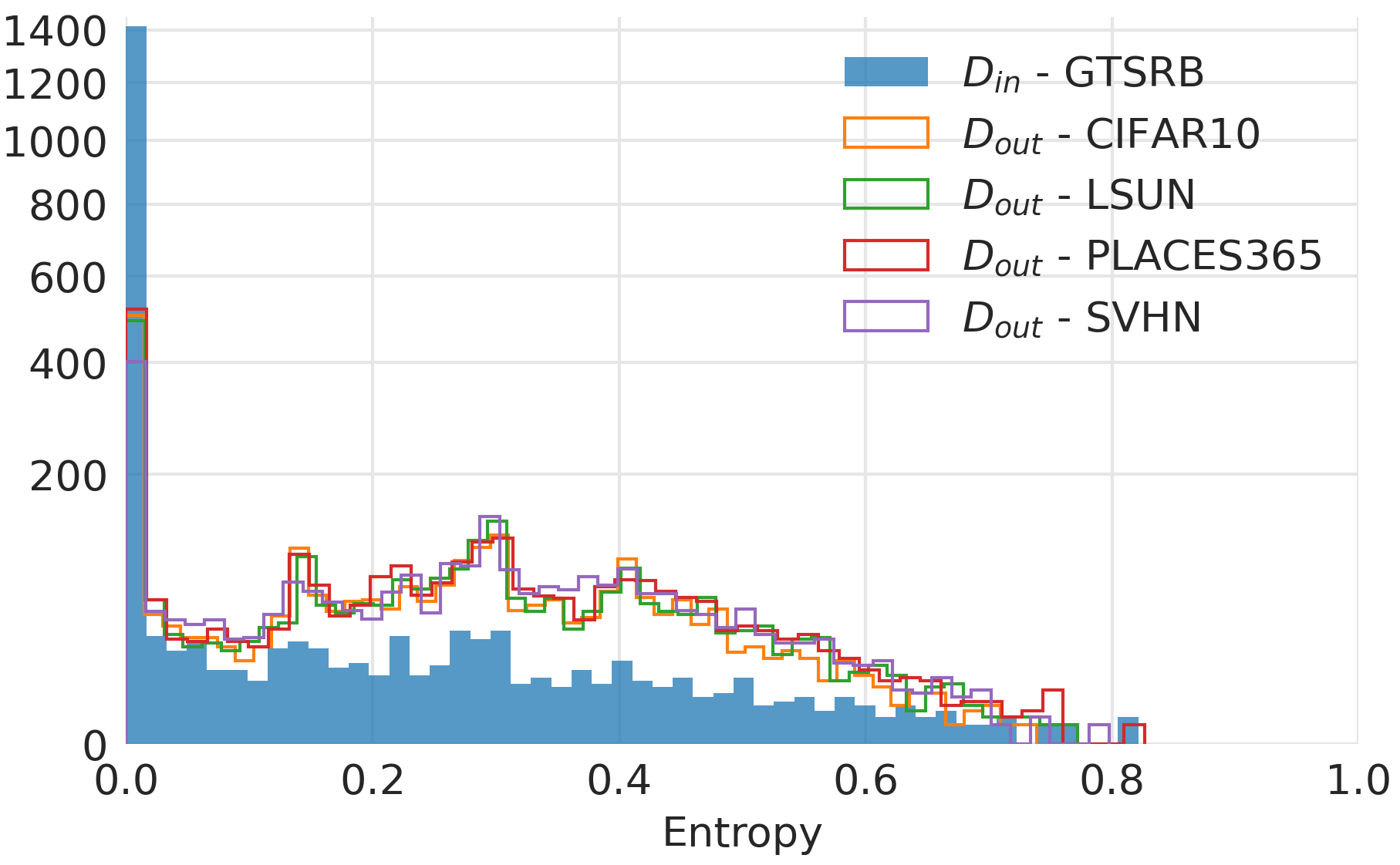}
  }%
  \caption{Comparison of entropy histograms between $D_{in}$ (GTSRB, filled blue bars) and several $D_{out}$ (not filled and coloured according to the dataset used) for different models. MCA-AE provides the best separation between $D_{in}$ and $D_{out}$. Notice the non-linear scale on the y-axis to improve visualization.}
  \label{fig:entropy}
\end{figure*}

We report the summary of our results for uncertainty and OOD detection in Table \ref{table:ood}. An interesting observation is the result that our approach performs significantly better when the visual complexity is increased (GTSRB, SVIRO), while the performance of MC Dropout and ensemble of models decreases on those setups. On the other side, on visually much simpler datasets (MNIST, Fashion-MNIST, SVHN) the performance of MC dropout and ensemble of models performs best. Thus, our method seems to be more beneficial for higher visual complexity, but this behavior should be investigated in detail in future work. Another interesting observation is that our approach provides better OOD estimations for the unseen Tesla vehicle from SVIRO. It can be observed that the different SVIRO-Uncertainty splits are much more challenging and undergo a large performance gap for all methods.

We computed the histograms of the entropies for each $D_{in}$ and $D_{out}$ and report the results in Fig. \ref{fig:entropy} when trained on GTSRB. The results show that the entropy distribution between $D_{in}$ and several $D_{out}$ are best separated by our approach. The distributions of the different $D_{out}$ are more similar then for the other models. To quantify this, we computed the sum of the Wasserstein distances between $D_{in}$ and all $D_{out}$ (TD, larger is better, as we want them to be different) separately and the sum of the distances between $D_{out}$ CIFAR10 and all other $D_{out}$ (OD, smaller is better, as we want them to be similar). We then computed the mean and standard deviation across 10 runs. The results are reported in Table \ref{table:ae-uncertainty-entropy} and show that our method best separates uncertainty between $D_{in}$ and $D_{out}$. Further, all $D_{out}$ are most similar between each other.

\begin{table}
  \small
  \caption{We computed the sum of the Wasserstein distances between $D_{in}$ and all $D_{out}$ (TD $\uparrow$) separately and the sum of the distances between $D_{out}$ CIFAR10 and all other $D_{out}$ (OD $\downarrow$) over 10 runs. We report mean and standard deviation.}
  \label{table:ae-uncertainty-entropy}
  \centering
  \begin{tabular}{rlccccc}
    \toprule
    & & MCA-AE (Ours) & MC Dropout & Ensemble\\
    \midrule
    OD & $\downarrow$ & {\cellcolor{good}} $0.049 \pm 0.007$  & $0.080 \pm 0.016$ & $0.050 \pm 0.007$ \\
    TD & $\uparrow$ & {\cellcolor{good}} $1.551 \pm 0.044$  & $0.854 \pm 0.028$ & $0.686 \pm 0.017$ \\
    \bottomrule
  \end{tabular}
\end{table}

\subsection{Ablation study}
We want to highlight that the performance of our method is improved due to the recursive application of the previously trained autoencoder. To this end we provide additional results where we compare the performance if no recursion is applied. We repeat the evaluation from the previous section and report the performance in Table \ref{table:ablation}. By comparing the results against Table \ref{table:ood}, it becomes apparent that the recursive application significantly improves uncertainty and OOD estimation.

\begin{table}
  \caption{OOD and uncertainty estimation when no recursion is applied. In most cases the results are worse compared to 2 recursions - see Table \ref{table:ood}. In case they are better, we mark them grey.}
  \label{table:ablation}
  \centering
  \begin{tabular}{l@{\quad}S[table-format=2.1(1)]@{\quad}S[table-format=2.1(1)]S[table-format=2.1(1)]}
    \toprule
    $\mathcal{D}_{in} \to \mathcal{D}_{out}$ & AUROC \hspace{3pt} $\uparrow$ & AUPR \hspace{3pt} $\uparrow$ & FPR\SI{95}{\percent} \hspace{3pt} $\downarrow$ \\
    \midrule
    MNIST $\to$ MNIST &       73.5 \pm 1.6 & 91.3 \pm 0.8 & 82.6 \pm 2.7 \\
    MNIST $\to$ CIFAR10 &     80.4 \pm 3.9 & 77.8 \pm 4.5 & 58.9 \pm 8.6 \\
    MNIST $\to$ Fashion &     61.5 \pm 5.8 & 59.3 \pm 5.4 & 82.7 \pm 4.2 \\
    MNIST $\to$ Omniglot &    32.6 \pm 10.5 & 44.0 \pm 6.5 & 99.9 \pm 0.1 \\
    MNIST $\to$ SVHN &        87.2 \pm 3.4 & 83.3 \pm 4.9 & 38.8 \pm 15.5 \\
    \midrule
    Fashion $\to$ Fashion &     77.2 \pm 0.9 & 91.6 \pm 0.5 & 82.0 \pm 1.7 \\
    Fashion $\to$ CIFAR10 &     88.2 \pm 5.0 & 89.6 \pm 6.2 & 63.0 \pm 15.9 \\
    Fashion $\to$ MNIST &       $\cellcolor{good}$ 88.2 \pm 3.3 & $\cellcolor{good}$ 89.3 \pm 3.0 & 55.7 \pm 8.9 \\
    Fashion $\to$ Omniglot &    60.9 \pm 25.4 & 71.3 \pm 19.5 & 98.5 \pm 2.8 \\
    Fashion $\to$ SVHN &        87.2 \pm 6.7 & 88.2 \pm 8.8 & 61.0 \pm 11.6 \\
    \midrule
    SVHN $\to$ SVHN &         67.2 \pm 1.2 & 54.6 \pm 2.3 & 87.9 \pm 2.1 \\
    SVHN $\to$ CIFAR10 &      56.0 \pm 1.0 & 57.2 \pm 1.1 & 94.8 \pm 0.9 \\
    SVHN $\to$ GTSRB &        53.6 \pm 3.0 & 60.4 \pm 2.7 & 94.9 \pm 1.9 \\
    SVHN $\to$ LSUN &         57.5 \pm 1.7 & 59.2 \pm 1.7 & 94.4 \pm 1.4 \\
    SVHN $\to$ Places365 &    57.9 \pm 1.3 & 58.6 \pm 1.4 & 93.9 \pm 1.5 \\
    \midrule
    GTSRB $\to$ GTSRB &        $\cellcolor{good}$ 85.7 \pm 1.3 & $\cellcolor{good}$ 95.9 \pm 0.6 & $\cellcolor{good}$ 67.3 \pm 2.9\\
    GTSRB $\to$ CIFAR10 &      82.2 \pm 2.3 & 81.0 \pm 2.5 & 69.9 \pm 6.5\\
    GTSRB $\to$ LSUN &         83.2 \pm 2.2 & 82.0 \pm 2.3 & 68.6 \pm 6.1\\
    GTSRB $\to$ Places365 &    82.8 \pm 2.1 & 81.3 \pm 2.3 & 68.2 \pm 5.9\\
    GTSRB $\to$ SVHN &         79.8 \pm 2.8 & 78.7 \pm 3.1 & 76.4 \pm 5.5\\
    \midrule
    SVIRO-U $\to$ CIFAR10 &    73.4 \pm 2.8 & 60.9 \pm 3.3 & 76.4 \pm 4.7\\
    SVIRO-U $\to$ GTSRB &      70.5 \pm 7.8 & 63.6 \pm 7.2 & 82.4 \pm 6.6\\
    SVIRO-U $\to$ LSUN &       70.8 \pm 2.7 & 58.2 \pm 2.6 & 81.2 \pm 3.6\\
    SVIRO-U $\to$ Places365 &  73.5 \pm 2.9 & 60.4 \pm 3.2 & 76.4 \pm 4.5 \\
    SVIRO-U $\to$ SVHN &       79.9 \pm 3.5 & 66.7 \pm 5.7 & 59.8 \pm 4.6\\
    \midrule
    SVIRO-U $\to$ Adults (A) &    86.7 \pm 2.2 & 98.6 \pm 0.5 & 66.6 \pm 8.8\\
    SVIRO-U $\to$ Seats (S) &     19.5 \pm 13.1 & 1.2 \pm 1.0 & $\cellcolor{good}$ 74.6 \pm 37.6\\
    SVIRO-U $\to$ Objects (O) &   58.6 \pm 4.9 & 56.6 \pm 6.2 & 88.2 \pm 6.2\\
    SVIRO-U $\to$ A,S &           43.4 \pm 2.9 & 11.0 \pm 2.0 & 95.1 \pm 2.3\\
    SVIRO-U $\to$ A,O &           65.9 \pm 1.8 & 75.1 \pm 1.6 & 88.5 \pm 1.4\\
    SVIRO-U $\to$ A,S,O &         48.4 \pm 3.2 & 16.9 \pm 2.1 & 91.6 \pm 2.4\\
    SVIRO-U $\to$ Tesla (OOD) &   54.2 \pm 10.8 & 86.2 \pm 4.1 & 94.8 \pm 3.7\\
    \bottomrule
  \end{tabular}
\end{table}

In Fig \ref{fig:reconstructions} we report the reconstructions after 1, 2, 3 and 4 iterative steps. We repeat this for models trained on different $D_{in}$ and show that $D_{out}$ reconstructions converge over time (and much slower) to training samples. We hence believe that considering the trajectory of the latent space representation over several steps can be an additional indicator whether an input sample is in- or out-of-distribution. It becomes also visible that the reconstruction converges robustly to similar classes for $D_{in}$ samples, but to different classes for $D_{out}$.

\section{Discussion and Limitations}
From a mathematical point of view dynamical systems are defined by natural phenomena or mechanical systems one wants to investigate and understand. Hence, designing or influencing the dynamical system of interest is usually not a possibility. An interesting observation is that the latter phenomenon is not the case for the recursive application of an autoencoder which is then interpreted as a dynamical system. Since we train the autoencoder in the first step, the resulting dynamical behavior and its attractors can be influenced by our previously defined autoencoder training procedure. We believe that it is an interesting direction for future work to analyze this interrelationship. Further, the effect of the number of epochs needed to obtain good results should be investigated. The basins of attraction can be studied after the autoencoder model is trained, such that potentially this information could be used to further improve robustness, interpretability and uncertainty estimation. We believe that the trajectory of the latent space representation over several iterations can give hints about the model robustness. Finally, while we fix the dropout mask for one recursion and each iterative step (but using a different one for each new recursion), it would also be possible to sample a new function $f$ for each iterative step within a recursion. 

\section{Conclusion}
Our results on several datasets show that the recursive application of autoencoder models, viewed as dynamical systems, together with an MC dropout approach provides good uncertainty and out-of-distributions estimations. Our model design choices improve the performance, particularly for computer vision datasets of higher visual complexity. Our ablation study highlights that the success is mainly due to the recursion and the entropy histograms underline the improved separability compared to MC dropout and an ensemble of models. 

\section*{Acknowledgement}
The first author is supported by the Luxembourg National Research Fund (FNR) under grant number 13043281. The second author is supported by DECODE (01IW21001). 



\bibliographystyle{IEEEtran}
\bibliography{bibliography}

\end{document}